\documentclass[11pt,a4paper]{amsproc}
\usepackage{times}
\usepackage{latexsym}

\usepackage[utf8]{inputenc}
\usepackage{tikz}
\usepackage{multirow}
\usepackage{geometry}
\usepackage{multicol}
\usepackage{microtype}

\title{Automated essay scoring using efficient transformer-based language models}
\author{Christopher M. Ormerod}
\address{Cambium Assessment, Inc.}
\curraddr{1000 Thomas Jefferson St., N.W. Washington, D.C. 20007}
\email{christopher.ormerod@cambiumassessment.com}
\author{Akanksha Malhotra}
\address{University of Colorado, Boulder}
\email{Akanksha.Malhotra@colorado.edu}
\author{Amir Jafari}
\address{Cambium Assessment, Inc.}
\curraddr{1000 Thomas Jefferson St., N.W. Washington, D.C. 20007}
\email{amir.jafari@cambiumassessment.com}
\date{\today}

\begin{document}

\begin{abstract}
Automated Essay Scoring (AES) is a cross-disciplinary effort involving Education, Linguistics, and Natural Language Processing (NLP). The efficacy of an NLP model in AES tests it ability to evaluate long-term dependencies and extrapolate meaning even when text is poorly written. Large pretrained transformer-based language models have dominated the current state-of-the-art in many NLP tasks, however, the computational requirements of these models make them expensive to deploy in practice. The goal of this paper is to challenge the paradigm in NLP that bigger is better when it comes to AES. To do this, we evaluate the performance of several fine-tuned pretrained NLP models with a modest number of parameters on an AES dataset. By ensembling our models, we achieve excellent results with fewer parameters than most pretrained transformer-based models.
\end{abstract}

\maketitle

\section{Introduction}

The idea that a computer could analyze writing style dates back to the work of Page in 1968 \cite{PEG}. Many engines in production today rely on explicitly defined hand-crafted features designed by experts to measure the intrinsic characteristics of writing \cite{OverviewAES}. These features are combined with frequency-based methods and statistical models to form a collection of methods that are broadly termed Bag-of-Word (BOW) methods \cite{BOW}. While BOW methods have been very successful from a purely statistical standpoint, \cite{babel} showed they tend to be brittle with respect novel uses of language and vulnerable to adversarially crafted inputs.

Neural Networks learn features implicitly rather than explicitly. It has been shown that initial neural network AES engines tend to be more more accurate and more robust to gaming than BOW methods \cite{AttAES, NNgaming}. In the broader NLP community, the recurrent neural network approaches used have been replaced by transformer-based approaches, like the Bidirectional Encoder Representations from Transformers (BERT) \cite{BERT}. These models tend to possess an order of magnitude more parameters than recurrent neural networks, but also boast state-of-the-art results in the General Language Understanding Evaluation (GLUE) benchmarks \cite{GLUE, GLUE2}. One of the main problems in deploying these models is their computational and memory requirements \cite{BERTAES}. This study explores the effectiveness of efficient versions of transformer-based models in the domain of AES. There are two aspects of AES that distinguishes it from GLUE tasks that might benefit from the efficiencies introduced in these models; firstly, the text being evaluated can be almost arbitrarily long. Secondly, essays written by students often contain many more spelling issues than would be present in typical GLUE tasks. It could be the case that fewer and more often updated parameters might possibly be better in this situation or more generally where smaller training sets are used. 

With respect to essay scoring the Automated Student Assessment Prize (ASAP) AES data-set on Kaggle is a standard openly accessible data-set by which we may evaluate the performance of a given AES model \cite{Kaggle}. Since the original test set is no longer available, we use the five-fold validation split  presented in \cite{NNAES} for a fair and accurate comparison. The accuracy of BERT and XLNet have been shown to be very solid on the Kaggle dataset \cite{OurBERT, Yang}. To our knowledge, combining BERT with hand-crafted features form the current state-of-the-art \cite{Uto}. 

The recent works of have challenged the paradigm that bigger models are necessarily better \cite{Electra, Reformer, PRADO, SHARNN}. The models in these papers possess some fundamental architectural characteristics that allow for a drastic reduction in model size, some of which may even be an advantage in AES. For this study, we consider the performance of the Albert models \cite{Albert}, Reformer models \cite{Reformer}, a version of the Electra model \cite{Electra}, and the Mobile-BERT model \cite{mobilebert} on the ASAP AES data-set. Not only are each of these models more efficient, we show that simple ensembles provide the best results to our knowledge on the ASAP AES dataset.

There are several reasons that this study is important. Firstly, the size of models scale quadratically with maximum length allowed, meaning that essays may be longer than the maximal length allowed by most pretrained transformer-based models. By considering efficiencies in underlying transformer architectures we can work on extending that maximum length. Secondly, as noted by \cite{BERTAES}, one of the barriers to effectively putting these models in production is the memory and size constraints of having fine-tuned models for every essay prompt. Lastly, we seek models that impose a smaller computational expense, which in turn has been linked by \cite{Energy} to a much smaller carbon footprint.

\section{Approaches to Automated Essay Scoring}\label{sec:AES}

From an assessment point of view, essay tests are useful in evaluating a student's ability to analyze and synthesize information, which assesses the upper levels of Blooms Taxonomy. Many modern rubrics use a multitude of scoring dimensions to evaluate an essay, including organization, elaboration, and writing style. An AES engine is a model that assigns scores to a piece of text closely approximating the way a person would score the text as a final score or in multiple dimensions. 

To evaluate the performance of an engine we use standard inter-rater reliability statistics. It has become standard practice in the development of a training set for an AES engine that each essay is evaluated by two different raters from which we may obtain a resolved score. The resolved score is usually either the same as the two raters in the case that they agree and an adjudicated score in cases in which they do not. In the case of the ASAP AES data-set, the resolved score for some items is taken to be the sum of the two raters, and a resolved score for others. The goal of a good model is to have a higher agreement with the resolved score in comparison with the agreement two raters have with each other. The most widely used statistic used to evaluate the agreement of two different collections of scores is the quadratic weighted kappa (QWK), defined by
\begin{equation}\label{eq:qwk}
\kappa = \frac{\sum \sum w_{ij} x_{ij}}{\sum \sum w_{ij} m_{ij}}
\end{equation}
where $x_{i,j}$ is the observed probability
\[
m_{i,j} = x_{ij}(1-x_{ij}),
\]
and
\[
w_{ij} = 1- \frac{(i-j)^2}{(k-1)^2},
\]
where $k$ is the number of classes. The other measurements used in the industry are the standardized mean difference (SMD) and the exact match or accuracy (Acc). The total number of essays and the human-human agreement for the two raters and the score ranges are shown in table \ref{tab:sum_hh}.

The usual protocol for training a statistical model is that some portion of the training set is set aside for evaluation while the remaining set is used for training. A portion of the training set is often isolated for purposes of early stopping and hyperparameter tuning. In evaluating the Kaggle dataset, we use the 5-fold cross validation splits defined by \cite{NNAES} so that our results are comparable to other works \cite{AESNN1, AttAES, OurBERT, Uto, Yang}. The resulting QWK is defined to be the average of the QWK values on each of the five different splits.

\begin{table*}[ht!]
    \centering
    \begin{tabular}{|p{3cm}|c | c| c| c|c|c|c|c|c|} \hline 
    & \multicolumn{8}{c|}{Essay Prompt}\\\hline
         &  1 & 2 & 3 & 4 & 5 & 6 & 7 & 8 \\\hline\hline
        Rater score range & 1-6 & 1-6& 0-3& 0-3& 0-4& 0-4 & 0-12 & 5-30\\ 
        Resolved score range & 2-12 & 1-6 & 0-3 & 0-3 & 0-4 & 0-4 & 2-24 & 10-60\\
        Average Length & 350& 350& 150& 150&150&150&250 & 650 \\ 
        Training examples & 1783 & 1800 & 1726 & 1772 & 1805 & 1800 & 1569 & 723 \\ \hline
        QWK & 0.721 & 0.814 & 0769 & 0.851 & 0.753 & 0.776 & 0.721 & 0.624 \\
        Acc & 65.3\% & 78.3\% & 74.9\% & 77.2\% & 58.0\% & 62.3\% & 29.2\% & 27.8\% \\
        SMD & 0.008 & 0.027 & 0.055 & 0.005 & 0.001 & 0.011 & 0.006 & 0.069\\\hline
    \end{tabular}
    \caption{A summary of the Automated Student Assessment Prize Automated Essay Scoring data-set.}
    \label{tab:sum_hh}
\end{table*}

\section{Methodology}\label{sec:Methodology}

Most engines currently in production rely on Latent Semantic Analysis or a multitude of hand-crafted features that measure style and prose. Once sufficiently many features are compiled, a traditional machine learning classifier, like logistic regression, is applied and fit to a training corpus. As a representative of this class of model we include the results of the ``Enhanced AI Scoring Engine", which is open source \footnote{https://github.com/edx/ease}. 

When modelling language with neural networks, the first layer of most neural networks are embedding layers, which send every token to an element of a semantic vector space. When training a neural network model from scratch, the word embedding vocabulary often comprises of the set of tokens that appear in the training set. There are several problems with this approach that come as a result of word sparsity in language and the presence of spelling mistakes. Alternatively, one may use a pretrained word embedding \cite{embedding} built from a large corpus with a large vocabulary with a sufficiently high dimensional semantic vector space. From an efficiency standpoint, these word embeddings alone can account for billions of parameters. We can shrink our embedding and address some issues arising from word sparsity by fixing a vocabulary of subwords using a version of the byte-pair-encoding (BPE) \cite{sentencepiece, BPE}. As such, pretrained models like BERT and those we use in this study utilize subwords to fix the size of the vocabulary. 

In addition to the word embeddings, positional embeddings and segment embedding are used to give the model information about the position of each word and next sentences prediction. The combination of the 3 embedding are the keys to reduce the vocabulary size, handling the out of vocab, and preserving the order of words.

Once we have applied the embedding layer to the tokens of a piece of text, it is essentially a sequence of elements of a vector space, which can be represented by a matrix whose dimensions are governed by the size of the semantic vector space and the number of tokens in the text.

In the field of language modeling, sequence-to-sequence models (seq2seq) in the form of this paper were proposed in \cite{seq2seq}. Initially, seq2seq models were used for natural machine translation between multiple languages. The seq2seq model has an encoder-decoder component; an encoder analyzes the input sequence and creates a context vector while the decoder is initialized with the context vector and is trainer to produce the transformed output. Previous language models were based on a fixed-length context vector and suffered from an inability to infer context over long sentences and text in general. An attention mechanism was used to improve the performance in translation for long sentences.

\begin{figure*}[!ht]
    \centering
    \begin{tikzpicture}[scale=1.0]
    \node[draw = black, thick, rectangle, rounded corners= 2mm, fill = blue!20](inp1) at (2,-1.5) {Embedding};
    \node[draw = black, thick, rectangle, rounded corners= 2mm, fill = blue!20](pos1) at (0,-.5) {Position};
    \node[circle, thick, draw=black](add1) at (2,-0.5) {$+$};
    \draw[thick,rounded corners= 4mm, fill = blue!5] (0,0.3) rectangle (3.5,5.5);
    \node[draw = black, thick, rectangle, rounded corners= 4mm, fill = blue!20](mh1) at (2,1.4) {\begin{tabular}{c}Multi-head\\ Attention\end{tabular}};
    \node[draw = black, thick, rectangle, rounded corners= 2mm, fill = blue!20](an1) at (2,2.4) {Add \& Norm};
    \node[draw = black, thick, rectangle, rounded corners= 4mm, fill = blue!20](ff1) at (2,4) {\begin{tabular}{c}Feed\\ Forward\end{tabular}};
    \node[draw = black, thick, rectangle, rounded corners= 2mm, fill = blue!20](an2) at (2,5) {Add \& Norm};
    \node[draw = black, thick, rectangle, rounded corners= 2mm, fill = blue!20](inp2) at (7,-1.5) {Embedding};
    \node[draw = black, thick, rectangle, rounded corners= 2mm, fill = blue!20](pos2) at (9,-.5) {Position};
    \node[circle, thick, draw=black](add2) at (7,-0.5) {$+$};
    \draw[thick,rounded corners= 4mm, fill = blue!5] (5,0) rectangle (8.5,8.5);
    \node[draw = black, thick, rectangle, rounded corners= 4mm, fill = blue!20](mmh1) at (7,1.4) {\begin{tabular}{c}Masked \\ Multi-head\\ Attention\end{tabular}};
    \node[draw = black, thick, rectangle, rounded corners= 2mm, fill = blue!20](an3) at (7,2.8) {Add \& Norm};
    \node[draw = black, thick, rectangle, rounded corners= 4mm, fill = blue!20](mh2) at (7,4.6) {\begin{tabular}{c}Multi-head\\ Attention\end{tabular}};
    \node[draw = black, thick, rectangle, rounded corners= 2mm, fill = blue!20](an4) at (7,5.6) {Add \& Norm};
    \node[draw = black, thick, rectangle, rounded corners= 4mm, fill = blue!20](ff2) at (7,7) {\begin{tabular}{c}Feed\\ Forward\end{tabular}};
    \node[draw = black, thick, rectangle, rounded corners= 2mm, fill = blue!20](an5) at (7,8) {Add \& Norm};
    \draw[thick,->] (pos1) -- (add1);
    \draw[thick,->] (inp1) -- (add1);
    \draw[thick,->] (add1) -- (mh1);
    
    \draw[thick,->] (pos2) -- (add2);
    \draw[thick,->] (inp2) -- (add2);
    \draw[thick,->] (add2) -- (mmh1);

    \draw[thick,->] (mh1) -- (an1);
    \draw[thick,->] (an1) -- (ff1);
    \draw[thick,->] (ff1) -- (an2);
    \draw[thick,->, rounded corners=4pt] (add1) -- (2,.5)--(0.5,.5) |- (an1);
    \draw[thick,->, rounded corners=4pt] (an1) -- (2,3) -- (.5,3) |- (an2);
    \draw[thick,->, rounded corners=4pt] (an2) -- (4,5) -- (4,3.5) -| (mh2);
    \draw[thick,->, rounded corners=4pt] (add2) -- (7,.3) -- (5.5,.3) |- (an3);
    \draw[thick,->, rounded corners=4pt] (mmh1) -- (an3);
    \draw[thick,->, rounded corners=4pt] (an3) -- (mh2);
    \draw[thick,->, rounded corners=4pt] (mh2) -- (an4);
    \draw[thick,->, rounded corners=4pt] (an4) -- (ff2);
    \draw[thick,->, rounded corners=4pt] (ff2) -- (an5);
    \draw[thick,->, rounded corners=4pt] (an3) -- (mh2);
    \node at (9,4) {$N\times $};
    \node at (-.5,4) {$N\times $};
    \end{tikzpicture}
    \caption{This is the basic architecture of a transformer-based model. The left block of $N$ layers is called the encoder while the right block of $N$ layers is called the decoder. The output of the Decoder is a sequence.}
    \label{fig:Transformer}
\end{figure*}
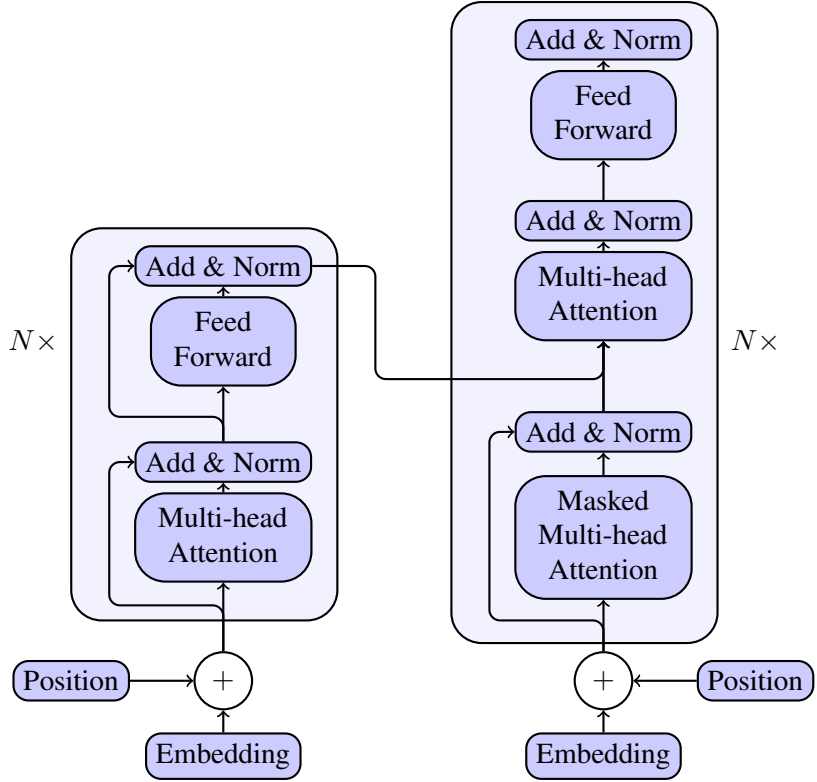

The use of a self-attention mechanism turns out to be very useful in th context of machine translation \cite{AttentionTranslation}. This mechanism, and it various derivatives, have been responsible for a large number of accuracy gains over a wide range of NLP tasks more broadly. The form of attention for this paper can be found in \cite{AttentionNeed}. Given a query matrix $Q$, key matrix $K$, and value matrix $V$, then the resulting sequence is given by
\begin{equation}\label{eq:attention}
Attention(Q,K,V) = \mathrm{softmax}\left(\frac{QK^T}{\sqrt{d_k}} \right)V.
\end{equation}
These matrices are obtained by linear transformations of either the output of a neural network, the output of a previous attention mechanism, or embeddings. The overall success of attention has led to the development of the transformer \cite{AttentionNeed}. The architecture of the transformer is outlined in Figure \ref{fig:Transformer}.

In the context of efficient models, if we consider all sequences up to length $L$ and if each query is of size $d$, then each keys are also of length $d$, hence, the matrix $QK^{T}$ of size $L\times L$. The implication is that the memory and computational power required to implement this mechanism scales quadratically with length. The above-mentioned models often adhere to a size restriction by letting $L = 512$. 

\section{Efficient Language Models}\label{sec:EffLM}

The transformer in language modeling is a innovative architecture to solve issues of seq2seq tasks while handling long term dependencies. It relies on self-attention mechanisms in its networks architecture. Self attention is in charge of managing the interdependence within the input elements.

In this study, we use five prominent models all of which are known to perform well as language models and possess an order of magnitude fewer parameters than BERT. It should be noted that many of the other models, like RoBERTa, XLNet, GPT models, T5, XLM, and even Distilled BERT all possess more than 60M parameters, hence, were excluded for this study. We only consider models utilizing an innovation that drastically reduces the number of parameters to at most one quarter the number of parameters of BERT. Of this class of models, we present a list of models and their respective sizes in Table \ref{tab:model_size}.

The backbone architecture of all above language models is BERT \cite{BERT}. It has become the state-of-the-art model for many different Natural Language Undestanding (NLU), Natural Language Generation (NLG) tasks including sequence and document classification.

\begin{table*}[!ht]
    \centering
    \begin{tabular}{||c||c | c |c ||} \hline \hline
        \bf Model & \bf Number   & \bf Training & \bf Inference\\
              & \bf of &  \bf Time    & \bf Time \\
              &\bf Parameters & \bf Speedup & \bf Speedup\\\hline \hline
        BERT (base) & 110M & $1.0\times$ & $1.0\times$ \\ \hline \hline
        Albert (base) & 12M & $1.0\times$ & $1.0 \times$\\
        Albert (large) & 18M & $0.7\times$ & $0.6 \times$\\
        Electra (small) & 14M & $3.8 \times$ & $2.4 \times$\\ 
        Mobile BERT & 24M & $2.5 \times$&  $1.7\times$ \\
        Reformer & 16M & $2.2\times$ & $1.6\times$ \\ \hline
        Electra + Mobile-BERT & 38M &$1.5\times $ & $1.0\times$\\ \hline
    \end{tabular}
    \caption{A summary of the memory and approximations of the computational requirements of the models we use in the study when comparing to BERT. These are estimates based on single epoch times using a fixed batch-size in training and inference.}
    \label{tab:model_size}
\end{table*}

The first models we consider are the Albert models of \cite{Albert}. The first efficiency of Albert is that the embedding matrix is factored. In the BERT model, the embedding size must be the same as the hidden layer size. Since the vocabulary of the BERT model is approximately 30,000 and the hidden layer size is 768 in the base model, the embedding alone requires approximately 25M parameters to define. Not only does this significantly increase the size of the model, we expect that some of those parameters to by updated rarely during the training process. By applying a linear layer after the embedding, we effectively factor the embedding matrix in a way that the actual embedding size can be much smaller than the feed forward layer. In the two models (large and base), the size of the vocabulary is about the same however the embedding dimension is effectively 128 making the embedding matrix one sixth the size.

A second efficiency proposed by Albert is that the layers share parameters. \cite{Albert} compare a multitude of parameter sharing scenarios in which all their parameters are shared across layers. The base and large Albert models, with 12 layer and 24 layers respectively, only possess a total of 12M and 18M parameters. The hidden size of of the base and large models are also 768 and 1024 respectively. Increasing the number of layers does increase the number of parameters but does come with a computational cost. In the pretraining of the Albert models, the models are trained with a sentence order prediction (SOP) loss function instead of next sentence prediction. It is argued by \cite{Albert} that SOP can solve NSP tasks while the converse is not true and that this distinction leads to consistent improvements in downstream tasks.

The second model we consider is the small version of Electra model presented by \cite{Electra}. Like the Albert model, there is a linear layer between the embedding and the hidden layers, allowing for a embedding size of 128, a hidden layer size of 256, and only four attention heads. The pretraining of the Electra model is trained as a pair of neural networks consisting of a generator and a discriminator with weight-sharing between the two networks. The generators role is to replace tokens in a sequence, and is therefore trained as a masked language model. The discriminator tries to identify that tokens were replaced by the generator in the sequence.

The third model, Mobile-BERT, was presented in \cite{mobilebert}. This model uses the same embedding factorization used to decouple the embedding size of 128 with the hidden size of 512. The main innovation of \cite{mobilebert} is that they decrease the model size by introducing a pair of linear transformations, called bottlenecks, around the transformer unit so that the transformer unit is operating on a hidden size of 128 instead of the full hidden size of of 512. This effectively shrinks the size of the underlying building blocks. MobileBERT uses absolute position embeddings and it is efficient at predicting masked tokens and at NLU.

The Reformer architecture of \cite{Reformer} differs from BERT most substantially by its version of an attention mechanism and that the feedforward component is of the attention layers use revsible residual layers. This means that, like \cite{Reversible}, the inputs of each layer can be computed on demand instead of being stored. \cite{Reformer} noted that they use an approximation of the (\ref{eq:attention}) in which the linear transformation used to define $Q$ and $K$ are the same, i.e., $Q=K$. When we calculate $QK^{T}$, more importantly, the softmax, we need only consider values in $Q$ and $K$ that are close. Using random vectors, we may create a Locally Sensitive Hashing (LSH) scheme, allowing us to chunk key/query vectors into finite collections of vectors that are known to contribute to the softmax. Each chunk may be computed in parallel changing the complexity from scaling with length as $O(L^2)$ to $O(L\log L)$. This is essentially a way to utilize the sparse nature of the attention matrix, attempting to not calculate pairs of values not likely to contribute to the softmax. More hashes improves the hashing scheme and better approximates the attention mechanism.

\section{Results}\label{sec:Results}

The networks above are all pretrained to be seq2seq models. While some of the pretraining differs for some models, as discussed above, we are required to modify a sequence-to-sequence neural network to produce a classification. Typically, the way in which this is done is that we take the first vector of the sequence as a finite set of features from which we may apply a classification. Applying a linear layer to these features produces a fixed length vector that is used for classification. 

\begin{table*}[!ht]
\begin{small}
\begin{tabular}{|p{3.3cm}| c|c | c | c | c | c | c | c| c |}\hline 
\bf QWK scores & \multicolumn{7}{c}{\bf Prompt}                                     && AVG  \\ \hline
                & \bf 1 & \bf 2 & \bf 3 & \bf 4 & \bf 5 & \bf 6 & \bf 7 & \bf 8 & QWK \\ \hline
EASE            & 0.781 & 0.621 & 0.630 & 0.749 & 0.782 & 0.771 & 0.727 & 0.534 &  0.699 \\
LSTM            & 0.775 & 0.687 & 0.683 & 0.795 & 0.818 & 0.813 & 0.805 & 0.594 & 0.746 \\
LSTM+CNN        & 0.821 & 0.688 & 0.694 & 0.805 & 0.807 & 0.819 & 0.808 & 0.644 & 0.761 \\ 
LSTM+CNN+Att    & 0.822 & 0.682 & 0.672 & 0.814 & 0.803 & 0.811 & 0.801 & 0.705 & 0.764 \\ \hline

BERT(base)      & 0.792 & 0.680 & 0.715 & 0.801 & 0.806 & 0.805 & 0.785 & 0.596 & 0.758\\
XLNet           & 0.777 & 0.681 & 0.693 & 0.806 & 0.783 & 0.794 & 0.787 & 0.627 & 0.743 \\ 
BERT + XLNet    & 0.808 & 0.697 & 0.703 & 0.819 & 0.808 & 0.815 & 0.807 & 0.605 & 0.758 \\ 
R${}^2$BERT & 0.817 & 0.719 & 0.698 & 0.845 & 0.841 & 0.847 & 0.839 & 0.744 & 0.794 \\ 
BERT + Features & 0.852 & 0.651 & 0.804 & 0.888 & 0.885 & 0.817 & 0.864 & 0.645 & {\bf 0.801} \\ \hline
Electra (small) & 0.816 & 0.664 & 0.682 & 0.792 & 0.792 & 0.787 & 0.827 & 0.715 & 0.759\\
Albert (base)   & 0.807 & 0.671 & 0.672 & 0.813 & 0.802 & 0.816 & 0.826 & 0.700 & 0.763\\
Albert (large)  & 0.801 & 0.676 & 0.668 & 0.810 & 0.805 & 0.807 & 0.832 & 0.700 & 0.763\\
Mobile-BERT     & 0.810 & 0.663 & 0.663 & 0.795 & 0.806 & 0.808 & 0.824 & 0.731 & 0.762\\
Reformer       & 0.802 & 0.651 & 0.670 & 0.754 & 0.771 & 0.762 & 0.747 & 0.548 & 0.713\\\hline
Electra+Mobile-BERT & 0.823 & 0.683 & 0.691 & 0.805 & 0.808 & 0.802 & 0.835 & 0.748 & 0.774 \\ \hline
Ensemble & 0.831 & 0.679 & 0.690 & 0.825 & 0.817 & 0.822 & 0.841 & 0.748 & {\bf 0.782} \\ \hline

\end{tabular}
\end{small}
\caption{The first set of agreement (QWK) statistics for each prompt. EASE, LSTM, LSTM+CNN, and LSTM+CNN+Att, were presented by \cite{NNAES} and \cite{AttAES}. The results of BERT, BERT extensions, and XLNET have been presented in \cite{OurBERT, Uto, Yang}.  The remaining rows are the results of this paper.}
\label{tab:HowGooddus}
\end{table*}

Given a set of $n$ possible scores, we divide the interval $[0,1]$ into $n$ even sub-intervals and map each score to the midpoint of those intervals. So in a similar manner to \cite{NNAES}, we treat this classification as a regression problem with a loss function of the mean squared error. This is slightly different from the standard multilabel classification using a Cross-entropy loss function often applied by default to transformer-based classification problems. Using the standard pretrained models with an untrained linear classification layer with a sigmoid activation function, we applied a small grid-search using two learning rates and two batch sizes for each model. The model performing the best on the test set was applied to validation.



For the reformer model, we pretrained our own 6-layer reformer model using a hidden layer size of 512, 4 hashing functions, and 4 attention heads. We used a cased sentencepiece tokenization consisting of 16,000 subwords and the model was trained with a maximum length of 1024 on a large corpus of essay texts from various grades on a single Nvidia RTX8000. This model addresses the length constraints other essay models struggle with on transformer-based architectures. 


We see that both Electra and Mobile-BERT show performance higher than BERT itself despite being smaller and faster. Given the extensive hyper-parameter tuning performed in \cite{OurBERT}, we can only speculate that any additional gains may be due to architectural differences. 

Lastly, we took our best models and simply averaged the outputs to obtain an ensembles whose output is in the interval $[0,1]$, then applied the same rounding transformation to obtain scores in the desired range. We highlight the ensemble of Mobile-BERT and Electra because on their own, this ensembled model provides a big increase in performance over BERT with approximately the same computational footprint on its own. 

\section{Discussion}

The goal of this paper was not to achieve state-of-the-art, but rather to show that we can achieve significant results within a very modest memory footprint and computational budget. We managed to exceed previous known results of BERT alone with approximately one third the parameters. Combining these models with R${}^2$ variants of \cite{Yang} or with features, as done in \cite{Uto}, would be interesting since they do not add little to the computational load of the system. 

There were noteworthy additions to the literature we did not consider for various reasons. The Longformer of \cite{Longformer} uses a sliding attention window in which the resulting self-attention mechanism scales linearly with length, however, the number of parameters of the pretrained Longformer models often coincided or exceeded those of the BERT model. Like the Reformer model of \cite{Reformer}, the Linformer of \cite{Linformer}, Sparse Transformer of \cite{Sparse}, and Performer model of \cite{Performer} exploit the observation that the attention matrix (given by the softmax) can be approximated by collection of low-rank matrices. Their different mechanisms for doing so means their complexity scales differently. The Projection Attention Networks for Document Classification On-Device (PRADO) model given by \cite{PRADO} seems promising, however, we did not have access to a version of it we could use. The SHARNN of \cite{SHARNN} also looks interesting, however, we found that the architecture was difficult to use for transfer learning. 

Transfer learning and language models improved the performance of document classification texts in natural language processing domain. We should note that all these models are pertained with a large corpus except for the Reformer model and we perform the fine-tuing process. Better training could improve upon the results of the Reformer model.

There are several reasons these directions in research are important; as we are seeing more and more attention given to power-efficient computing for usage on small-devices, we think the deep learning community will see a greater emphasis on smaller efficient models in the future. Secondly, this work provides a stepping stone to the classifying and evaluating documents where the necessity for context to extend beyond the limits that current pretrained transformer-based language models allow. Lastly we think combining the approaches that seem to have a beneficial effect on performance should give smaller, better, and more environmentally friendly models in the future. 

\section*{Acknowledgments}

We would like to acknowledge Susan Lottridge and Balaji Kodeswaran for their support of this project.


\end{document}